\documentclass{article}
\usepackage{amsmath,epsfig,subfigure}
\usepackage{booktabs}
\usepackage{multirow}
\usepackage[preprint]{spconfa4}

\let\OLDthebibliography\thebibliography
\renewcommand\thebibliography[1]{
  \OLDthebibliography{#1}
  \setlength{\parskip}{0pt}
  \setlength{\itemsep}{0pt plus 0.3ex}
}

\begin{document}\sloppy

\def\x{{\mathbf x}}
\def\L{{\cal L}}

\title{Perceptual Quality Assessment for Digital Human Heads}
%
\name{Zicheng Zhang$^{1,2}$, Yingjie Zhou$^{1,2}$, Wei Sun$^{1,2}$, Xiongkuo Min$^{1,2}$, Yuzhe Wu$^{3}$, and Guangtao Zhai$^{1,2}$}
\address{$^{1}$Institute of Image Communication and Network Engineering, Shanghai Jiao Tong University, China \\
$^{2}$ Peng Cheng Laboratory, China \\
$^{3}$College of Information Science \& Technology, Donghua University, China \\
zzc1998@sjtu.edu.cn}

\maketitle

%
\begin{abstract}
Digital humans are attracting more and more research interest during the last decade, the generation, representation, rendering, and animation of which have been put into large amounts of effort. However, the quality assessment of digital humans has fallen behind. Therefore, to tackle the challenge of digital human quality assessment issues, we propose the first large-scale quality assessment database for three-dimensional (3D) scanned digital human heads (DHHs). The constructed database consists of 55 reference DHHs and 1,540 distorted DHHs along with the subjective perceptual ratings. Then, a simple yet effective full-reference (FR) projection-based method is proposed to evaluate the visual quality of DHHs. The pretrained Swin Transformer tiny is employed for hierarchical feature extraction and the multi-head attention module is utilized for feature fusion. The experimental results reveal that the proposed method exhibits state-of-the-art performance among the mainstream FR metrics. The database is released at https://github.com/zzc-1998/DHHQA.
\end{abstract}

\let\thefootnote\relax\footnotetext{This work was supported in part by NSFC (No.62225112, No.61831015), the Fundamental Research Funds for the Central Universities, National Key R\&D Program of China 2021YFE0206700, Shanghai Municipal Science and Technology Major Project (2021SHZDZX0102), and NSFC No.62101325. }

\section{Introduction}
\label{sec:intro}

Digital humans are digital simulations and models of human beings on computers, which have been widely employed in applications such as games, automobiles, metaverse, etc. Current research scopes are mainly focused on the generation, representation, rendering, and animation of digital humans \cite{zhu2019applications}. However, with the rapid development of VR/AR technologies, more and more viewers have a higher demand on visual quality of digital humans, which makes it necessary to carry out quality assessment studies on digital humans. Unfortunately, the collection of digital human models is very expensive and time-consuming, which requires the assistance of professional three-dimensional (3D) scanning devices and human subjects. Moreover, large-scale quality assessment subjective experiments on the visual quality of digital humans are not available currently.  Therefore, we conduct both subjective and objective quality assessment studies for digital human heads in this paper and we hope this study can promote the development of digital human quality assessment (DHQA).

In this paper, we first propose a large-scale quality assessment database for digital human heads (DHHs). A total of 255 human subjects are invited to participate in the DHH scanning experiment. We carefully select 55 high-quality generated DHH models as the reference stimuli (the DHH models are in the format of textured meshes), which cover both male and female, young and old subjects. Then we manually degrade the reference DHHs with 7 types of distortions including surface simplification, position compression, UV compression, texture sub-sampling, texture compression, color noise, and geometry noise, and obtain 1,540 distorted DHHs in total. A well-controlled subjective experiment is conducted to collect mean opinion scores (MOS) for distorted DHHs. Consequently, we obtain the largest subject-rate quality assessment database for digital human heads (DHHQA). We validate several mainstream quality assessment metrics for the benchmark. Further, we specifically design a deep neural network (DNN) based full-reference (FR) quality assessment method to boost the performance on DHH quality evaluation. The experimental results show that the proposed method has a better correlation with the subjective ratings.

\begin{table*}[!htp]
\centering
\renewcommand\arraystretch{0.5}
\caption{The comparison of 3D-QA databases and our database.}
\begin{tabular}{ccccc}
\toprule
Database        & Source   & Subject-rated Models  &Content   & Description            \\
\midrule
SJTU-PCQA \cite{sjtu-pcqa}    &10 &420  & Colored Point Cloud  & Humans, Statues          \\
WPC \cite{liu2022perceptual}   &20 &740  &  Colored Point Cloud   & Fruit, Vegetables, Tools           \\
LSPCQA \cite{liu2022point} &104 &1,240    &Colored Point Cloud   & Animals, Humans, Vehicles, Daily objects  \\
CMDM \cite{cmdm}  &5 & 80  & Colored Mesh   &   Humans, Animals, Statues  \\
TMQA \cite{nehme2022textured} &55 &3,000 & Textured Mesh & Statues, Animals, Daily objects\\
DHHQA(Ours)           & 55  & 1,540 & Textured Mesh  & Scanned Real Human Heads\\
\bottomrule
\end{tabular}
\vspace{-0.4cm}
\label{tab:comparison}
\end{table*}

\section{Related Works}
\subsection{Digital Human}
Sensing, Modeling, and driving digital humans have been popular research topics in computer vision and computer graphics. Namely, 3D human pose and shape estimation \cite{martinez2017simple,pavllo20193d} are of great significance for the representation of digital human models. Digital human animation generation \cite{liu2019ntu,shahroudy2016ntu} aims to drive digital human motion activities with text or audio. However, little attention is paid to the visual quality assessment of digital humans.

\vspace{-0.1cm}
\subsection{3D Quality Assessment}
\vspace{-0.1cm}
Currently, 3D quality assessment (3D-QA) mainly focuses on point cloud quality assessment (PCQA) and mesh quality assessment (MQA). To cope with the challenge of visual quality assessment of 3D models, substantial efforts have been made to carry out 3D quality assessment databases \cite{sjtu-pcqa,liu2022perceptual,liu2022point,cmdm,nehme2022textured} and the detailed comparison with the proposed database is listed in Table \ref{tab:comparison}, from which we can find that the proposed DHHQA database is the first large-scale database specially designed for DHHs.

The objective quality assessment methods for 3D-QA can be generally categorized into two types: model-based methods \cite{liu2022perceptual,liu2022point,cmdm,zhang2021no,zhang2022no} and projection-based methods \cite{sjtu-pcqa,nehme2022textured,fan2022no,zhang2022treating,zhang2022mm}. Model-based methods extract features directly from 3D models while projection-based methods infer the visual quality of 3D models from the rendered projections. Although the model-based methods are invariant to viewpoints, it is difficult to efficiently extract quality-aware features from 3D models since high-quality 3D models usually contain large numbers of dense points/vertices. While the projection-based methods can make use of the mature image/video quality assessment (IQA/VQA) models, therefore gaining better performance.


\begin{figure}[tbp]
\centering
\subfigure[]{
\begin{minipage}[t]{0.23\linewidth}
\centering
\includegraphics[width=1.6cm,height = 2cm]{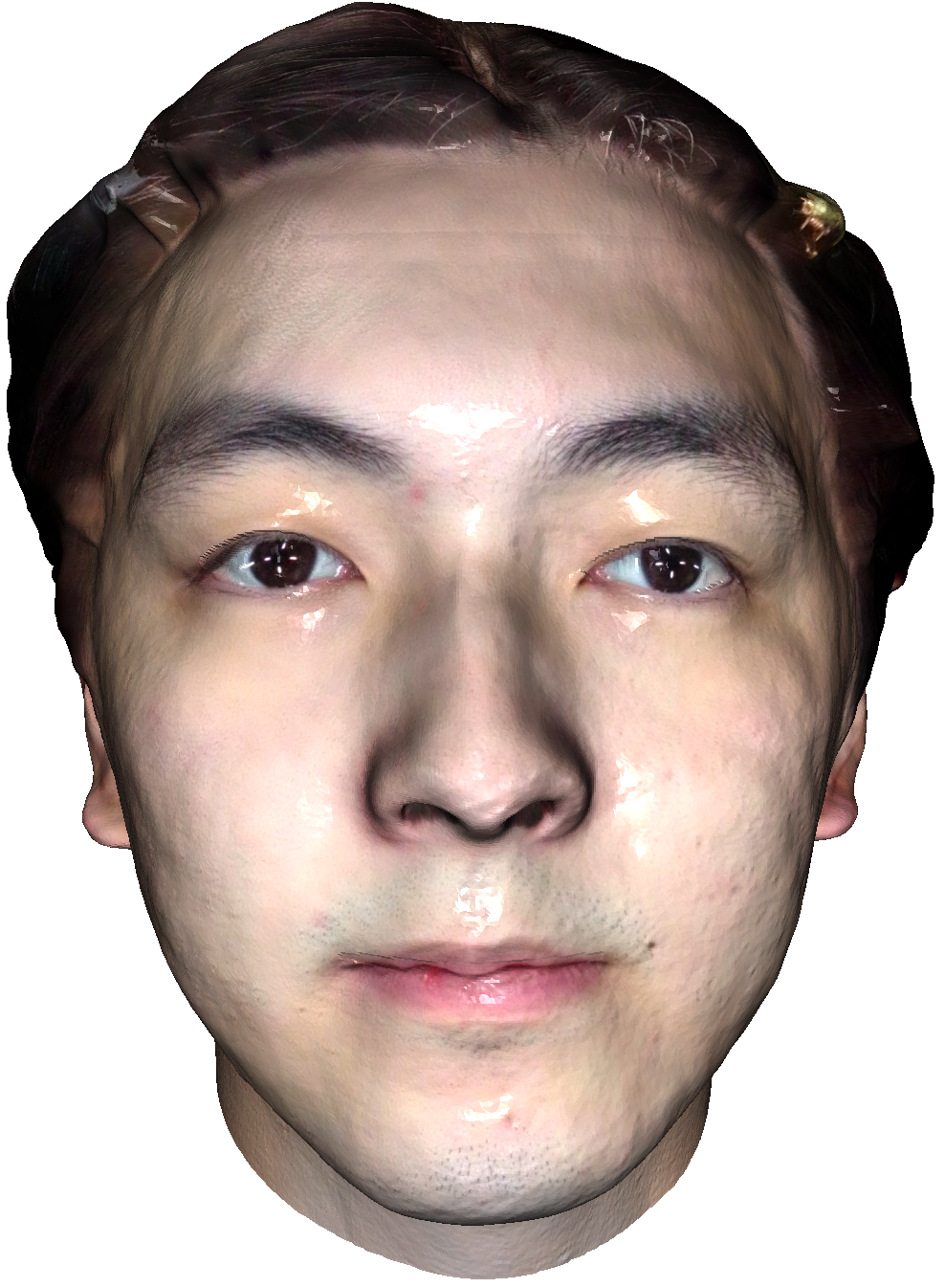}
\end{minipage}%
}%
\subfigure[]{
\begin{minipage}[t]{0.23\linewidth}
\centering
\includegraphics[width=1.5cm,height = 2cm]{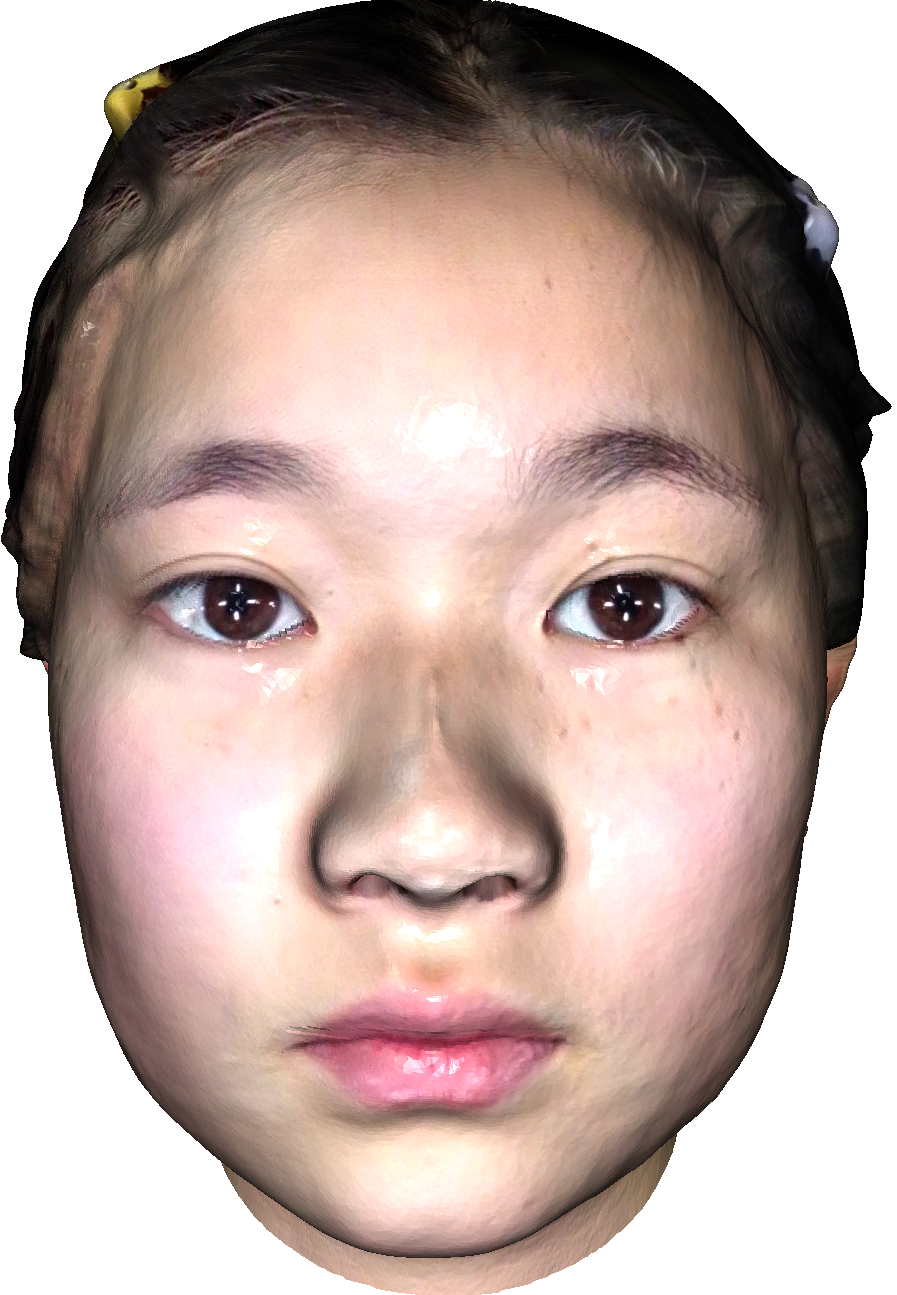}
\end{minipage}%
}%
\subfigure[]{
\begin{minipage}[t]{0.23\linewidth}
\centering
\includegraphics[width=1.6cm,height = 2.1cm]{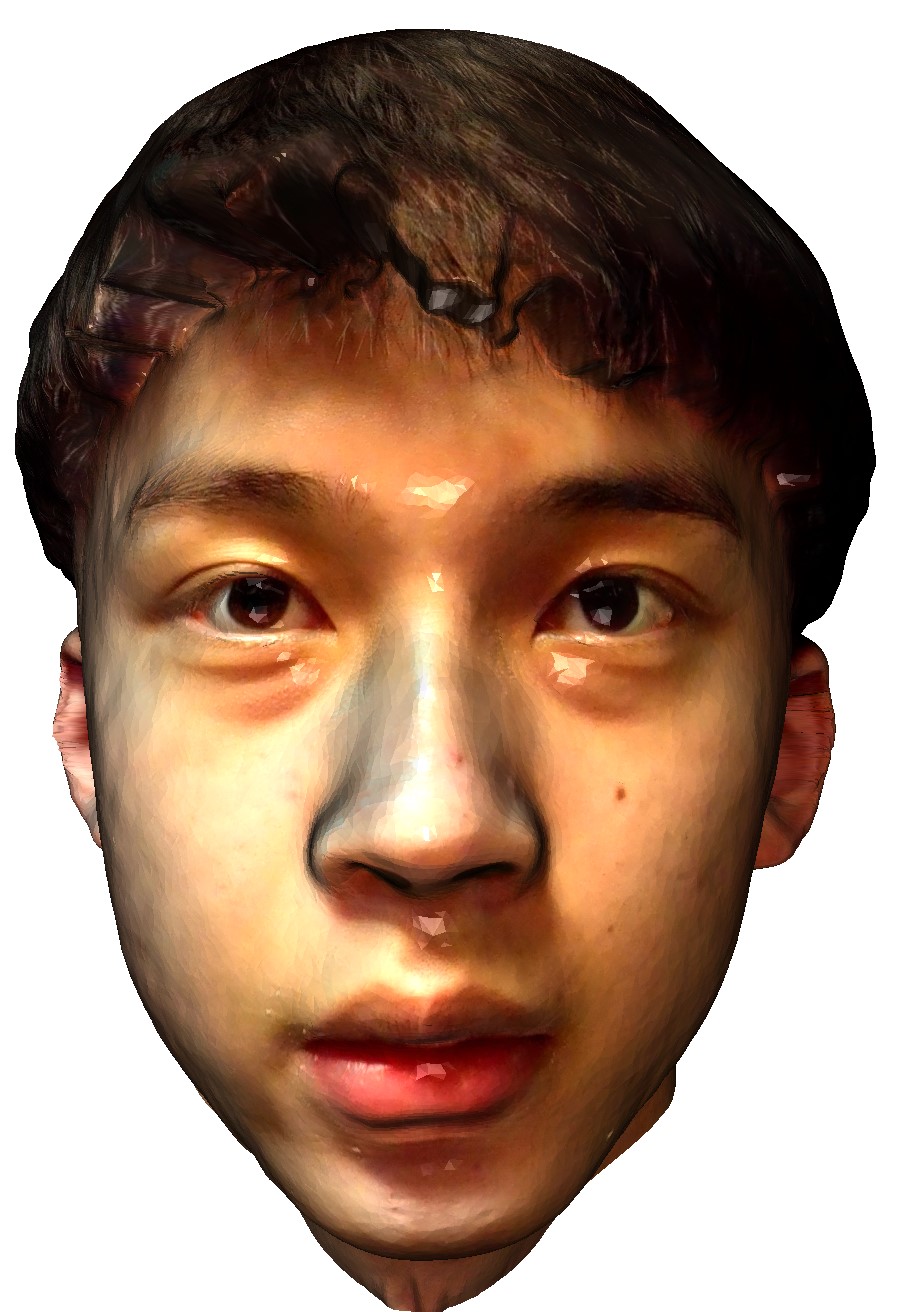}
\end{minipage}
}%
\subfigure[]{
\begin{minipage}[t]{0.23\linewidth}
\centering
\includegraphics[width=1.6cm,height = 2cm]{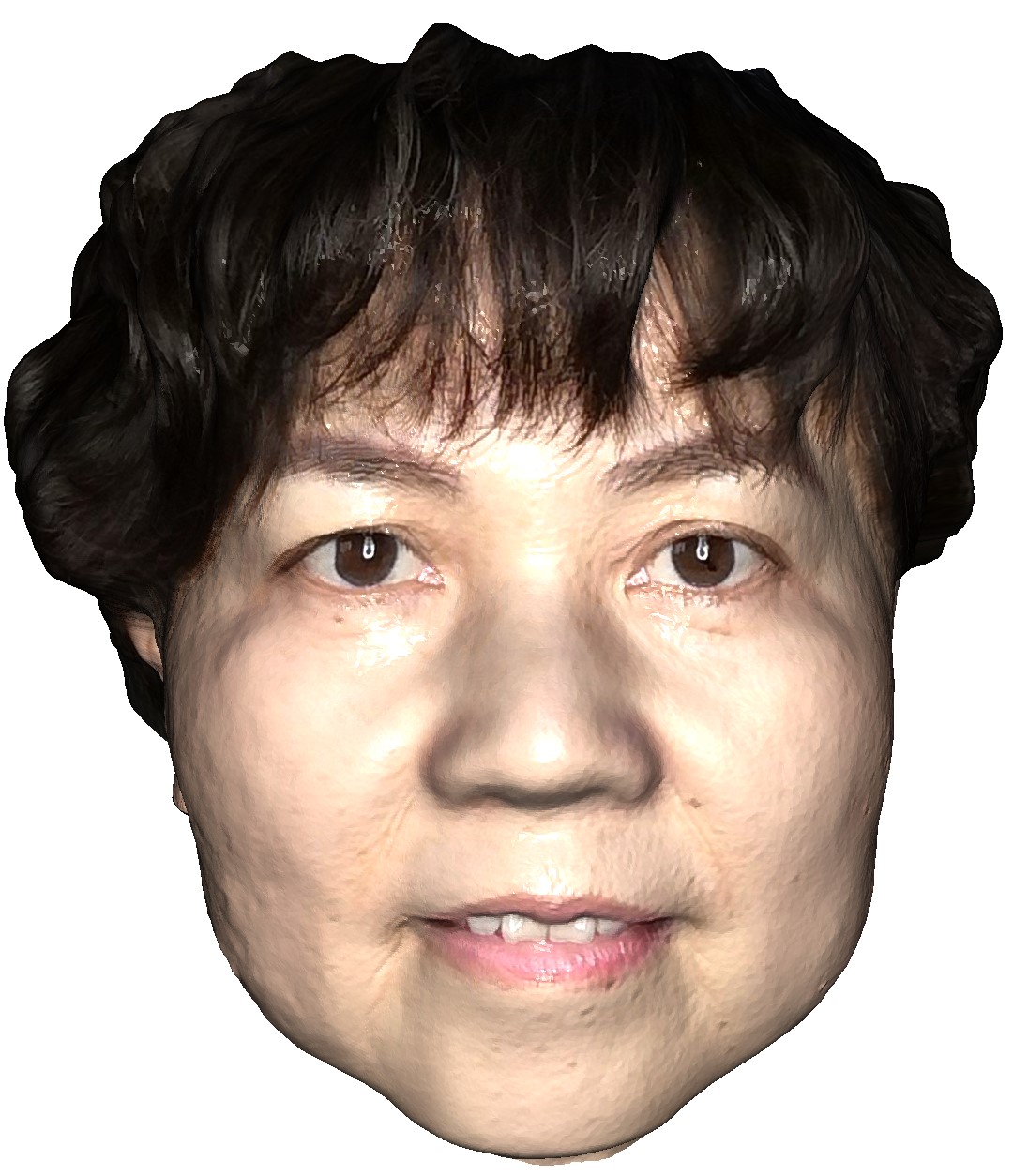}
\end{minipage}
}%

\centering
\vspace{-0.1cm}
\caption{DHH samples from the DHHQA database.}
\label{fig:sample}
\end{figure}

\vspace{-0.1cm}
\section{Database Construction}
\vspace{-0.1cm}
\subsection{DHH Collection \& Distortion Generation}
\vspace{-0.1cm}
\label{sec:distortion}
To gather the source models of DHHs, we invite 255 human subjects (aged from 12 to 60) to participate in the scanning experiment with the Bellus3D \footnote{https://www.bellus3d.com} app. Finally, a total of 55 DHHs are manually chosen as the source DHH models. Samples of the selected DHHs are exhibited in Fig. \ref{fig:sample}. 
Then 7 types of common distortions are applied to degrade the reference DHH models: 
(a) Surface simplification (SS): The algorithm proposed in \cite{garland1998simplifying} is employed to simplify the DHH models and the simplification rate is set as \{0.4, 0.2, 0.1, 0.05\}; (b) Position Compression (PC): The Draco \footnote{https://github.com/google/draco} library is applied to quantize the position attributes of the DHH models with compression parameter {\bf qp} set as \{6, 7, 8, 9\}; (c) UV Map Compression (UMC): The Draco library is applied to quantize the texture coordinate attributes with the compression parameter {\bf qt} set as \{7,8,9,10\}; (d) Texture Down-sampling (TD): The reference texture maps (4096$\times$4096) are down-sampled with resolutions of \{2048$\times$2048, 1024$\times$1024, 512$\times$512, 256$\times$256\}; (e) Texture Compression (TC): The JPEG compression is used to compress the texture maps with quality levels set as \{3, 10, 15, 20\}; (f) Geometry Noise (GN): Gaussian noise is introduced to the vertices of the DHH models with $\sigma_{g}$ set as \{0.05, 0.1, 0.15, 0.2\}; (g) Color Noise (CN): Gaussian noise is added to the texture maps with $\sigma_{c}$ set as \{20, 40, 60, 80\}. To sum up, 1,540 = 55$\times$7$\times$4 distorted DHH models are generated and the degradation levels are manually defined to cover the most quality range.

\begin{figure}[!tbp]
\centering

\subfigure[SS]{
\begin{minipage}[t]{0.23\linewidth}
\centering
\includegraphics[width=1.5cm,height = 2cm]{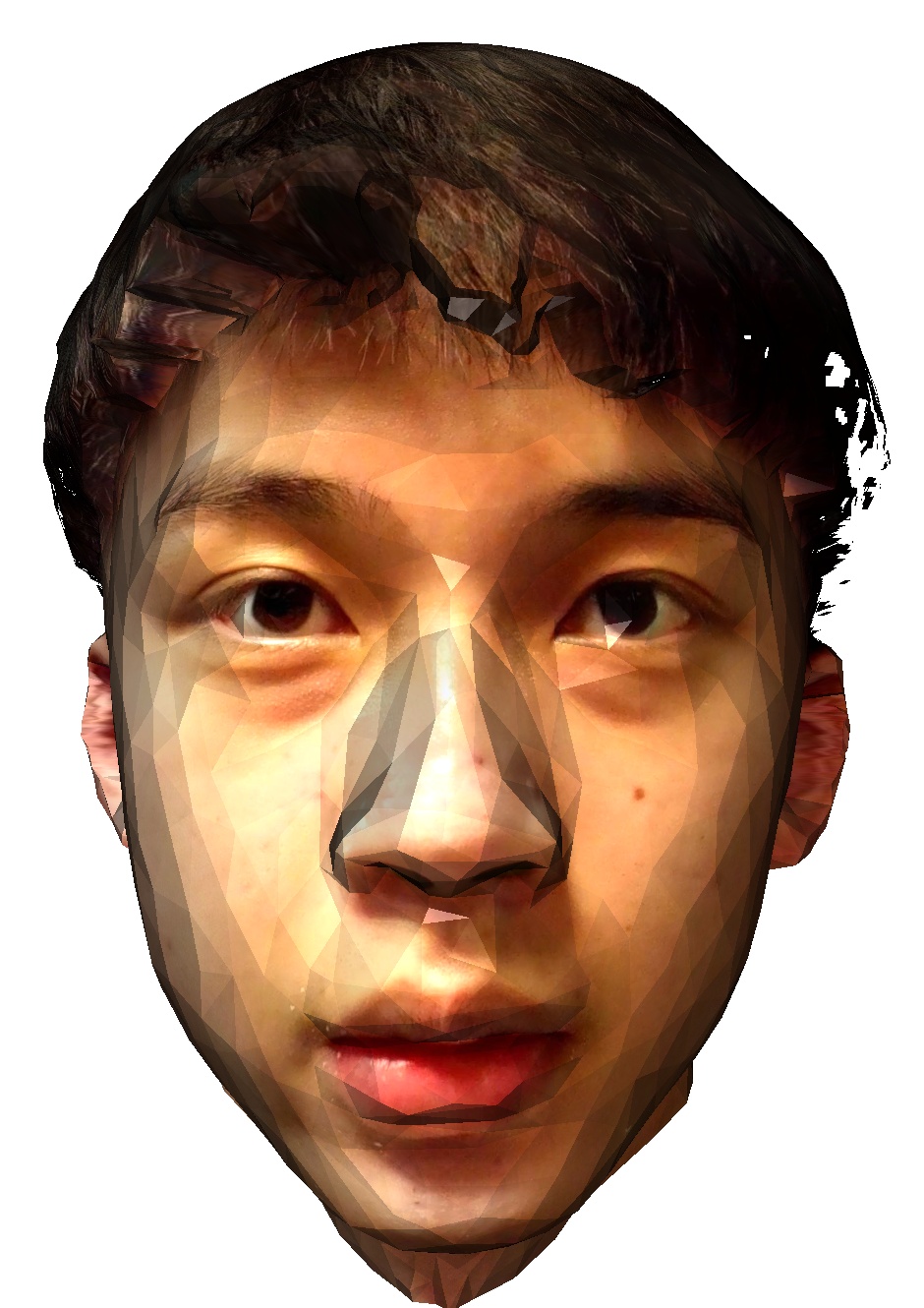}
\end{minipage}%
}%
\subfigure[PC]{
\begin{minipage}[t]{0.23\linewidth}
\centering
\includegraphics[width=1.5cm,height = 2cm]{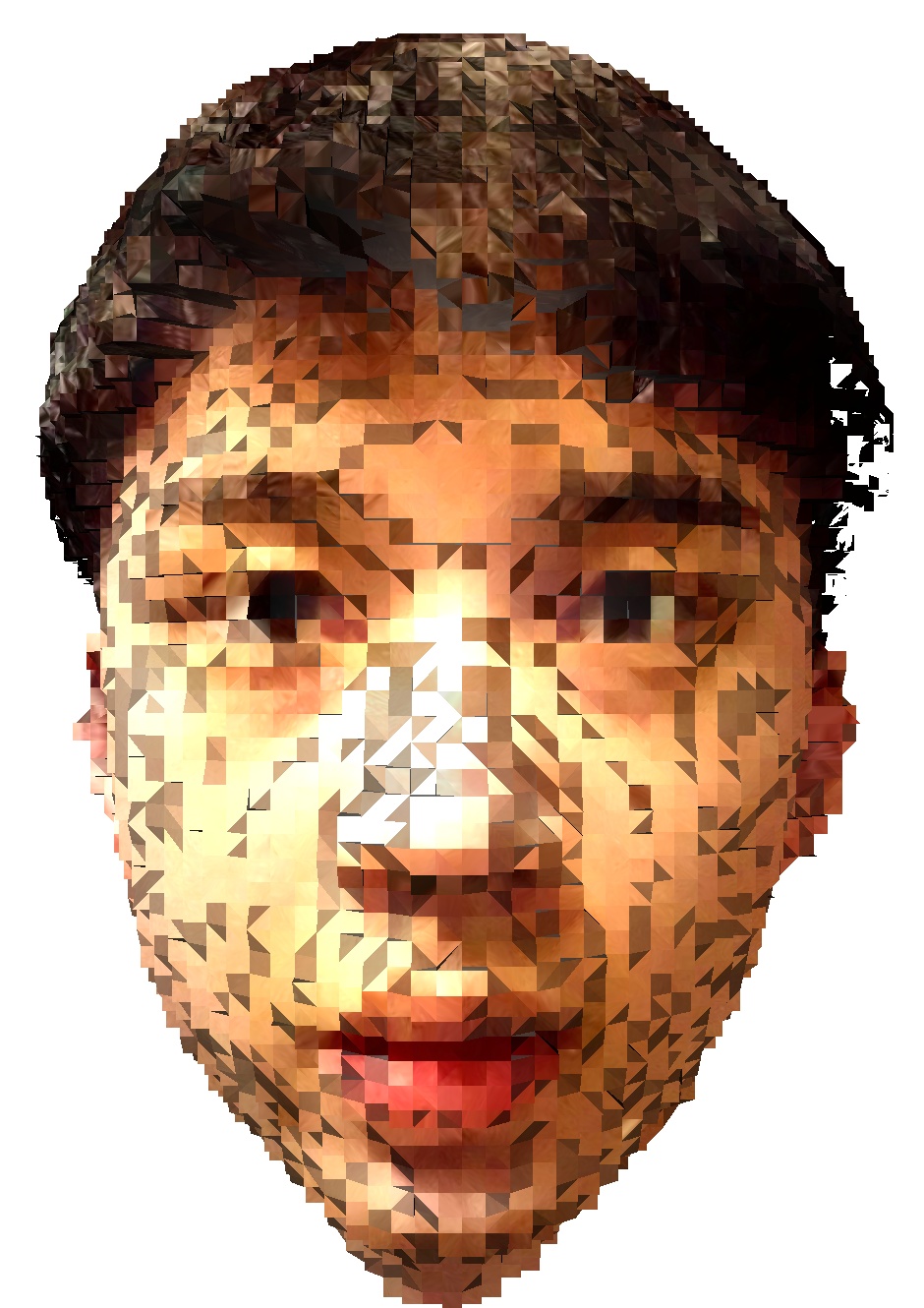}
\end{minipage}%
}%
\subfigure[UMC]{
\begin{minipage}[t]{0.23\linewidth}
\centering
\includegraphics[width=1.5cm,height = 2cm]{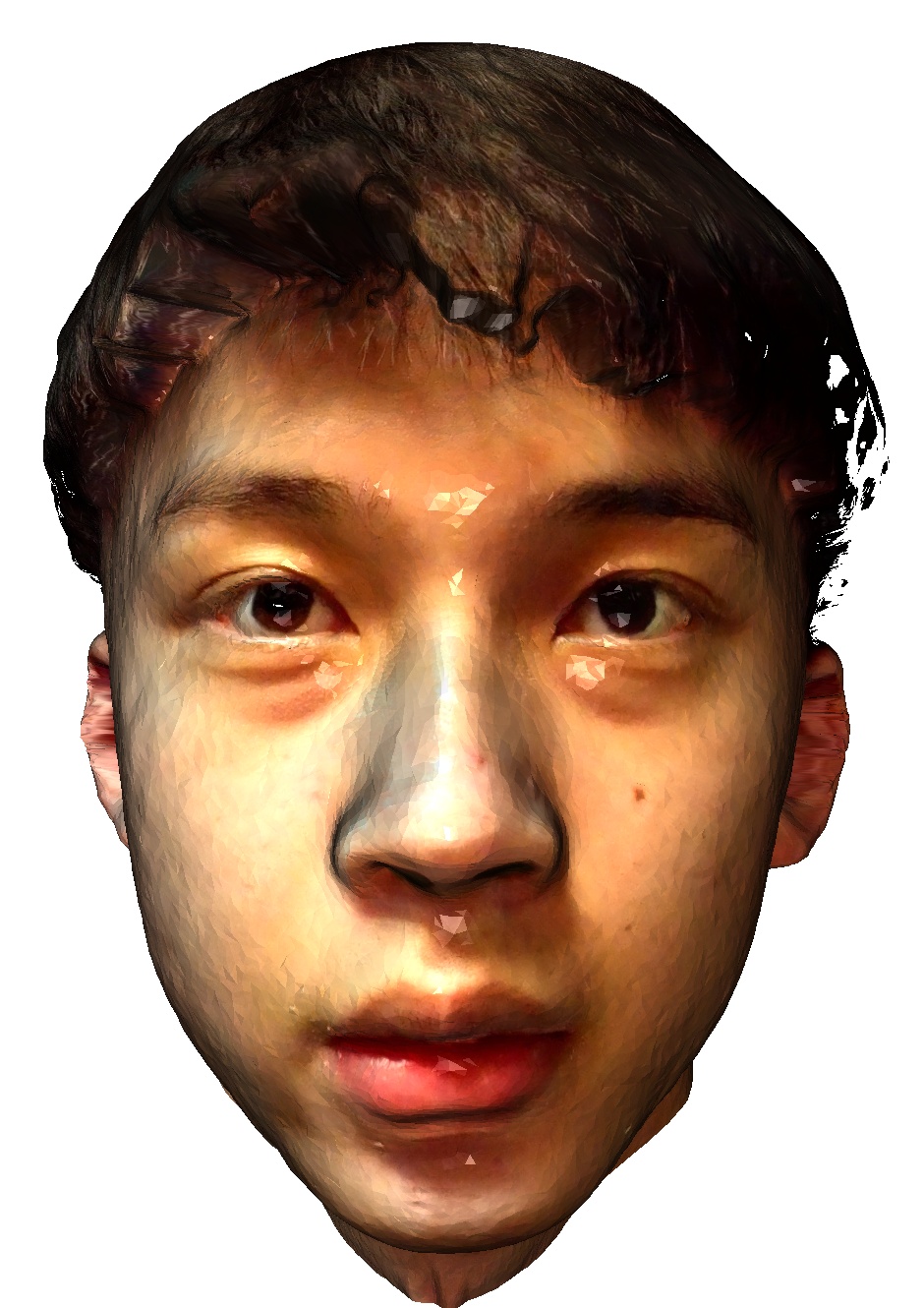}
\end{minipage}
}%
\subfigure[TD]{
\begin{minipage}[t]{0.23\linewidth}
\centering
\includegraphics[width=1.5cm,height = 2cm]{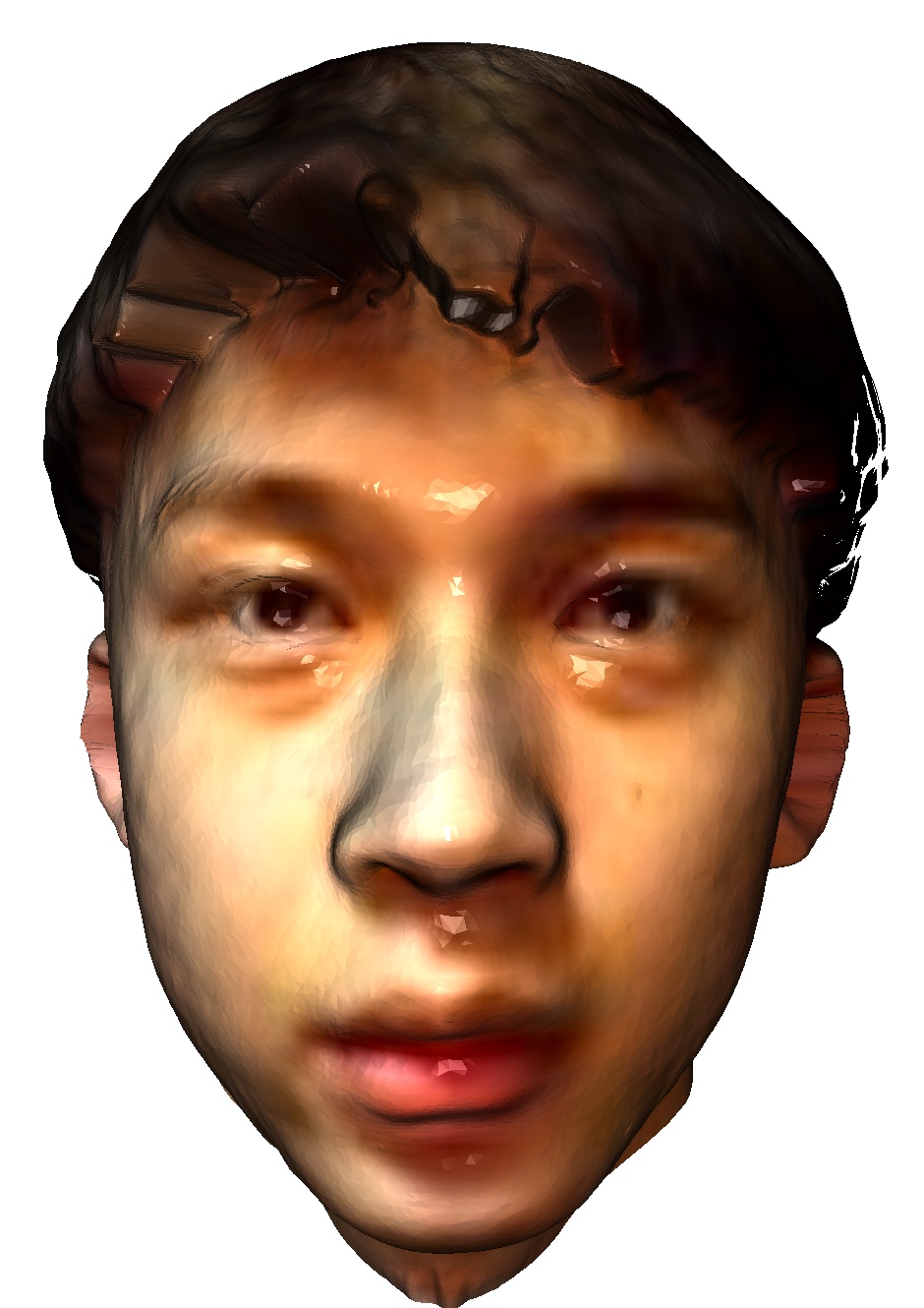}
\end{minipage}
}%
\\
\vspace{-0.1cm}
\subfigure[TC]{
\begin{minipage}[t]{0.23\linewidth}
\centering
\includegraphics[width=1.5cm,height = 2cm]{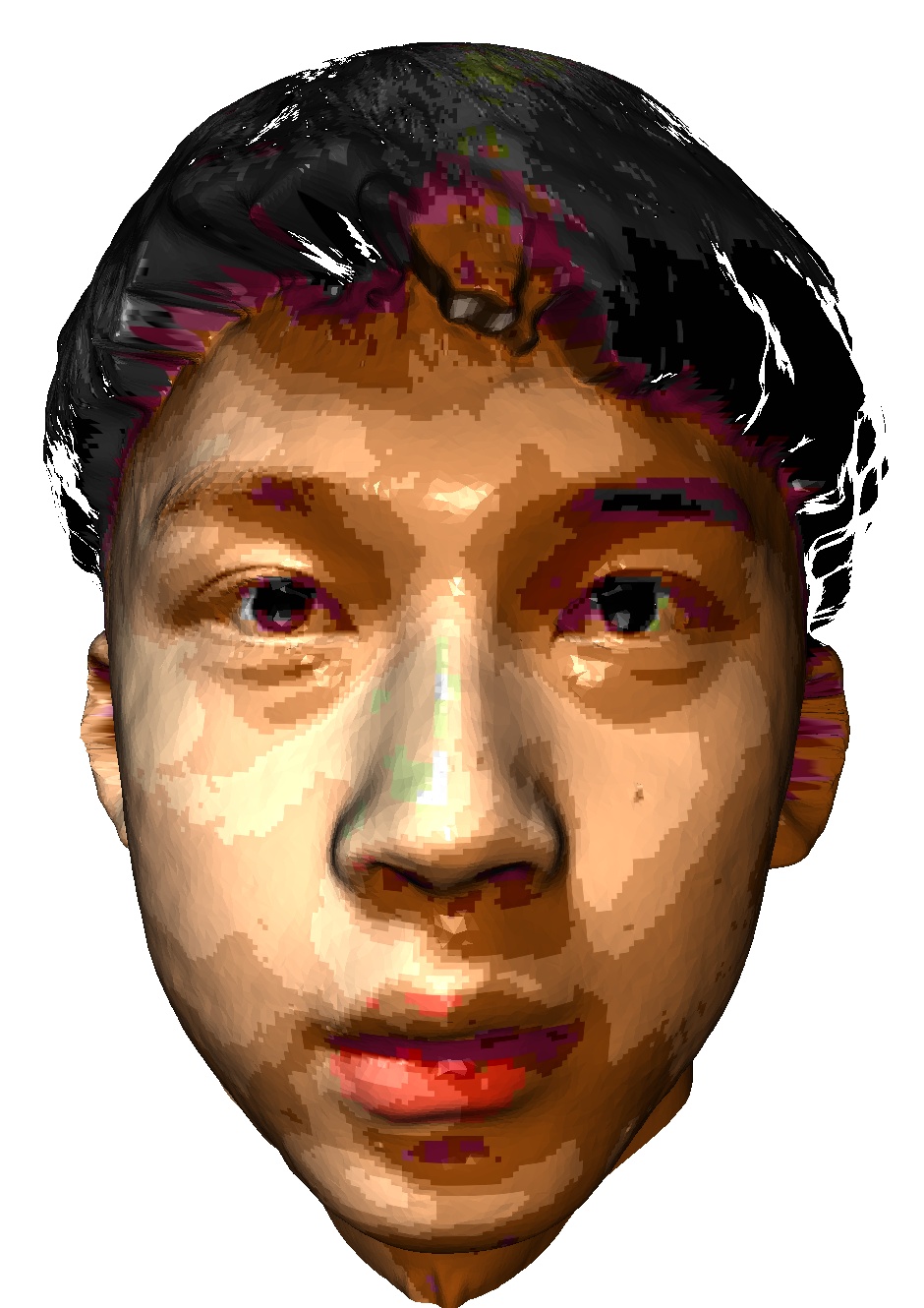}
\end{minipage}
}%
\subfigure[GN]{
\begin{minipage}[t]{0.23\linewidth}
\centering
\includegraphics[width=1.5cm,height = 2cm]{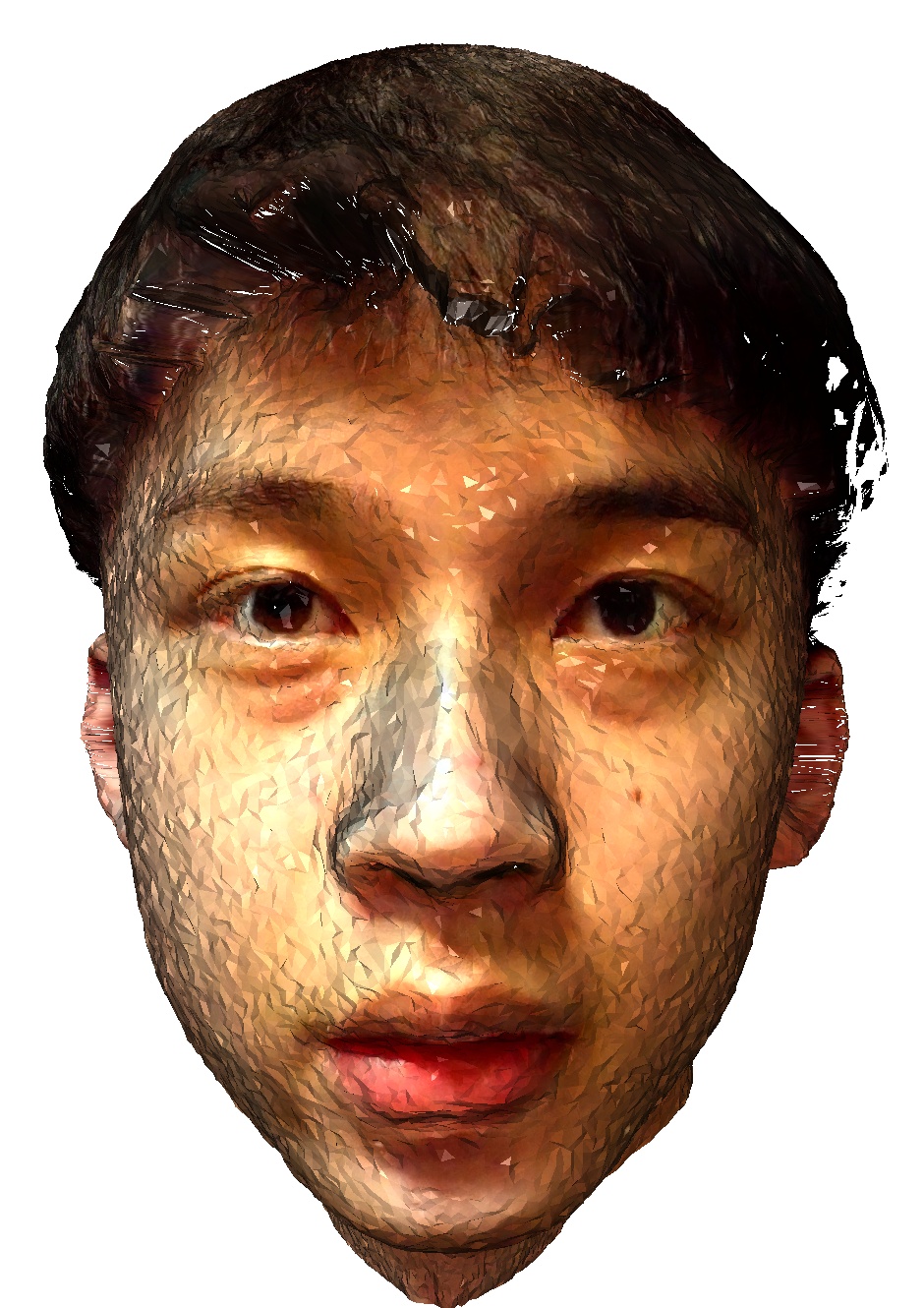}
\end{minipage}
}%
\subfigure[CN]{
\begin{minipage}[t]{0.23\linewidth}
\centering
\includegraphics[width=1.5cm,height = 2cm]{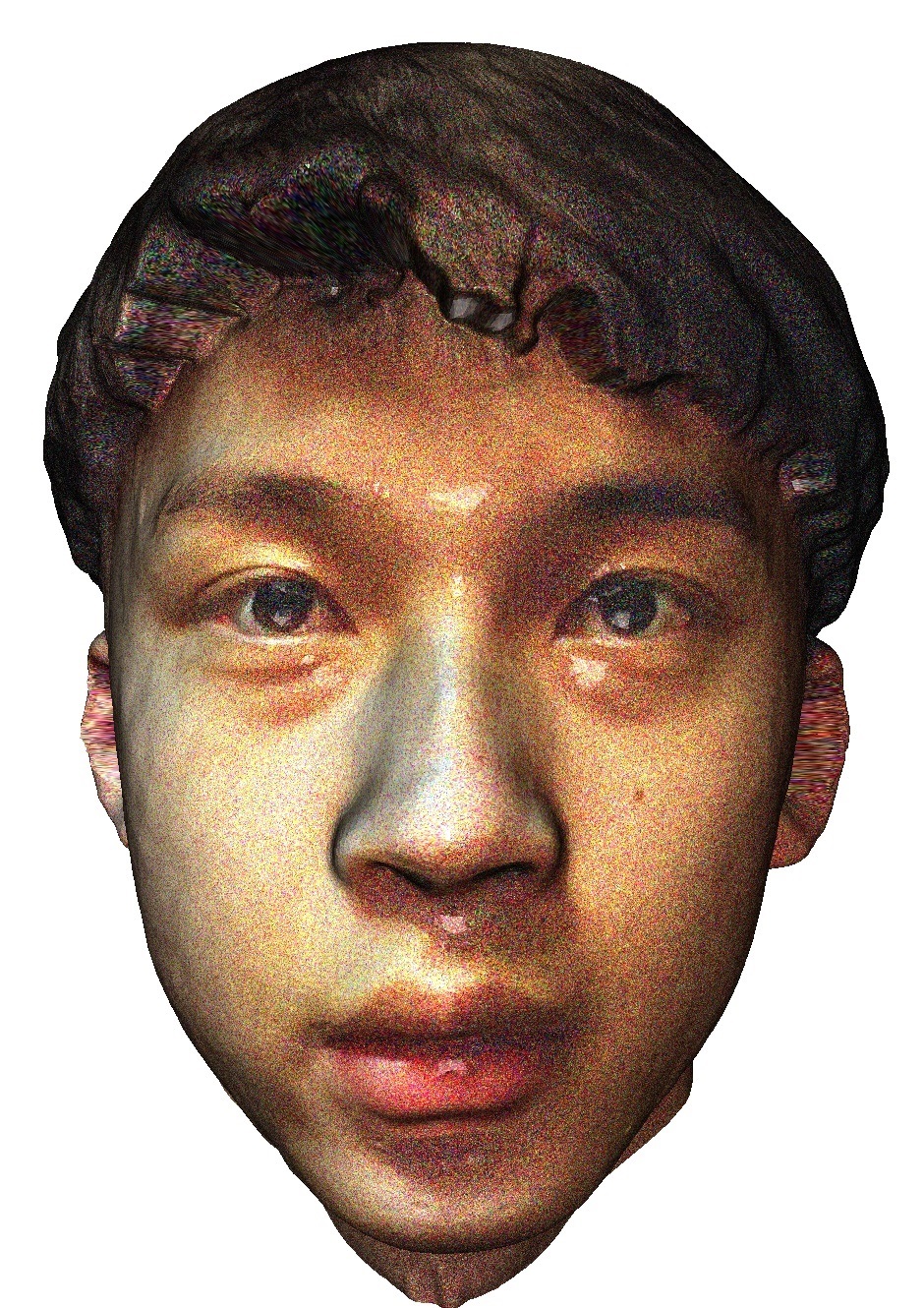}
\end{minipage}
}%
\subfigure[Ref]{
\begin{minipage}[t]{0.23\linewidth}
\centering
\includegraphics[width=1.5cm,height = 2cm]{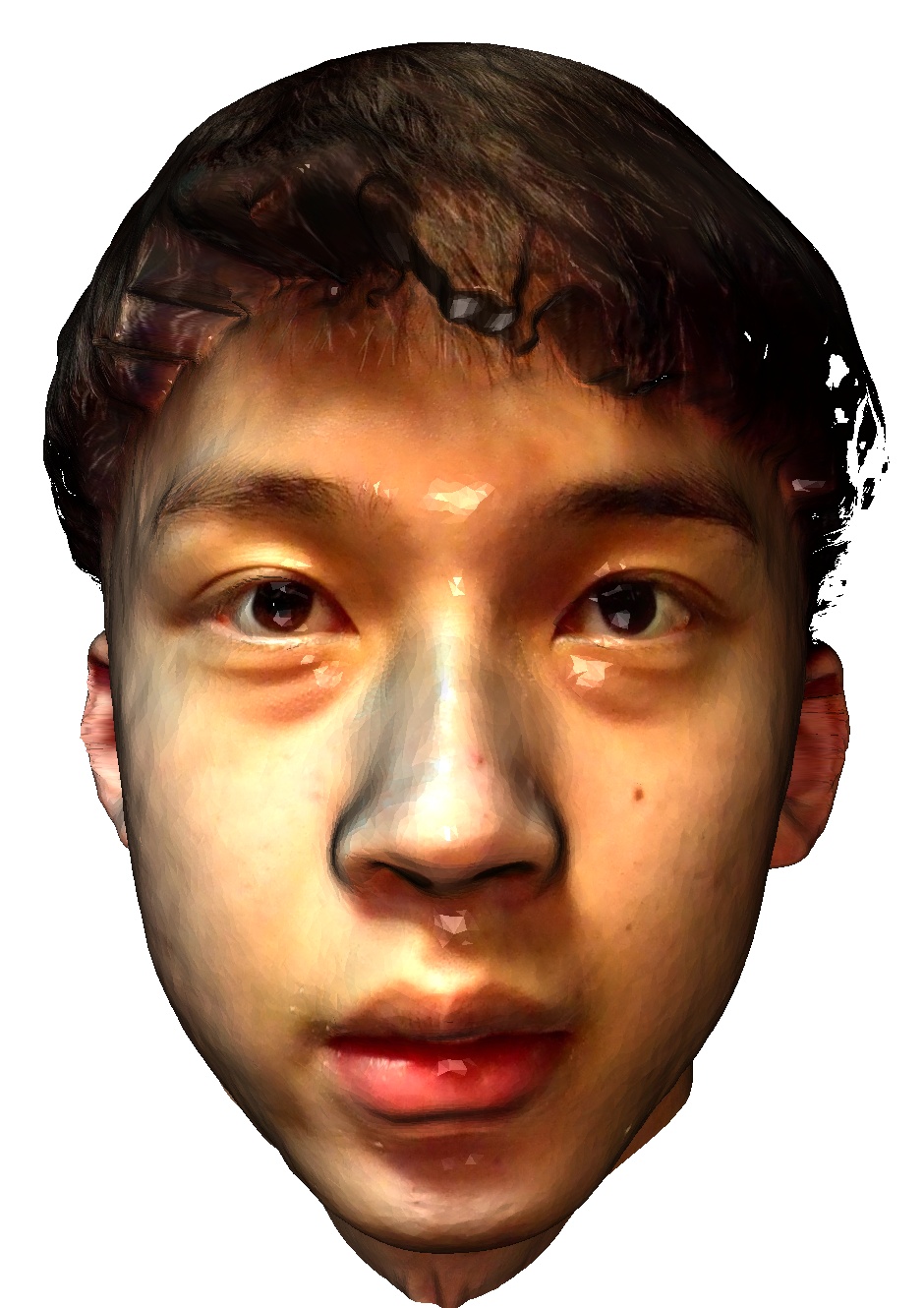}
\end{minipage}%
}%

\centering
\caption{Examples of DHH models with different types of distortions.}
\label{fig:distortion}
\vspace{-0.2cm}
\end{figure}

\vspace{-0.1cm}
\subsection{Subjective Quality Assessment Experiment}
\begin{figure}[!tp]
    \centering
    \includegraphics[width=6.4cm]{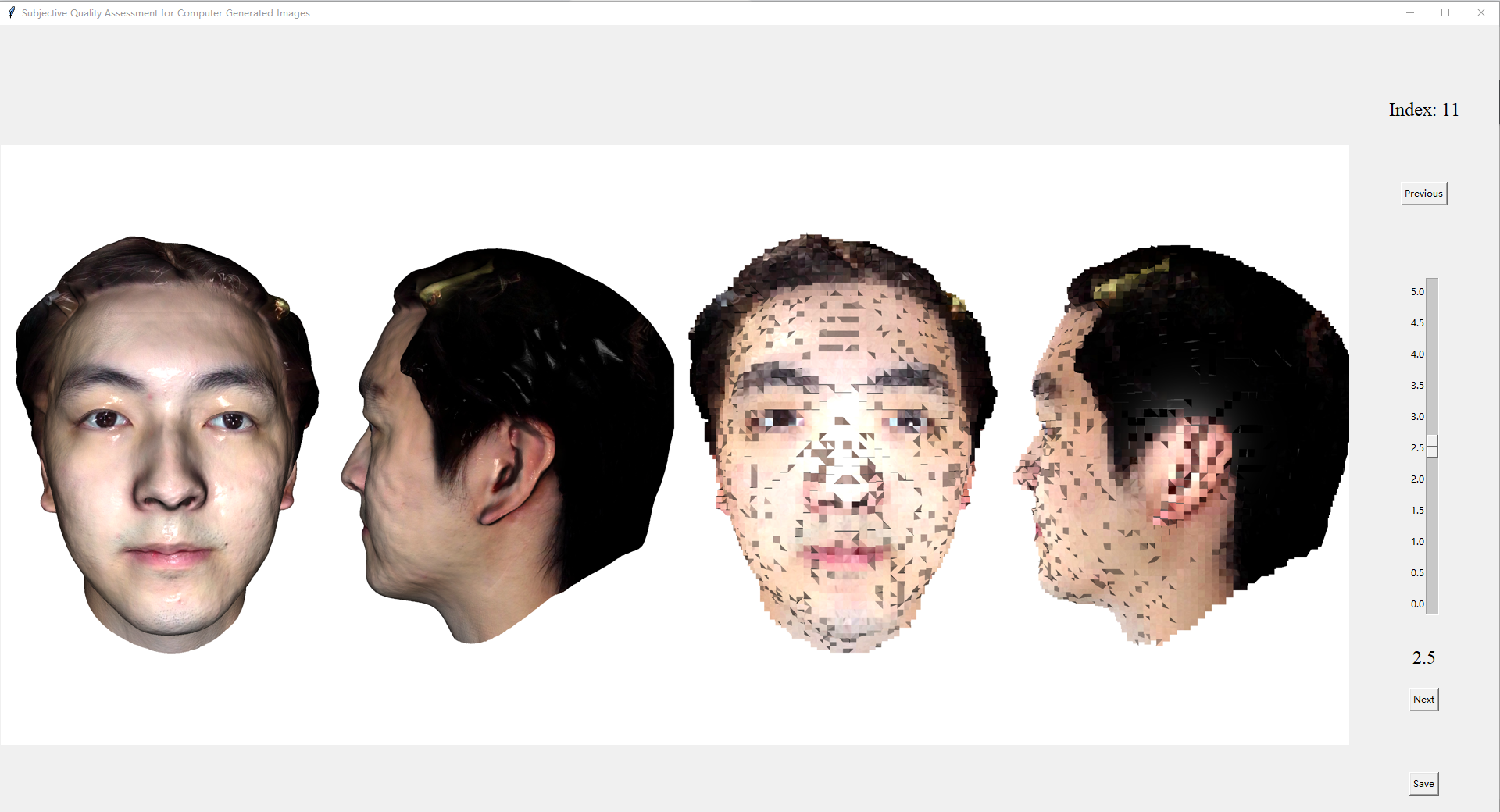}
    \vspace{-0.2cm}
    \caption{Illustration of the rating interface.}
    \label{fig:interface}
    \vspace{-0.2cm}
\end{figure}

\vspace{-0.1cm}
Following the recommendation of ITU-R BT.500-13 \cite{bt2002methodology}, we conduct the subjective quality assessment experiment in a well-controlled laboratory environment. All the DHH models are rendered into projections from the front and left side viewpoint and the reference as well as distorted projections are at the same time for evaluation. The projections are shown in random order with an interface designed by Python Tkinter on an iMac monitor which supports the resolution up to 4096 $\times$ 2304, the screenshot of which is illustrated in Fig. \ref{fig:interface}. 22 males and 18 females are invited to participate in the subjective experiment. The whole experiment is divided into 11 sessions. Each session contains 140 distorted DHH models and is attended by at least 20 participants. In all, more than 30,800 = 1,540$\times$20 subjective ratings are collected.

\begin{figure}[!t]
    \centering
    \includegraphics[width = 6.5cm]{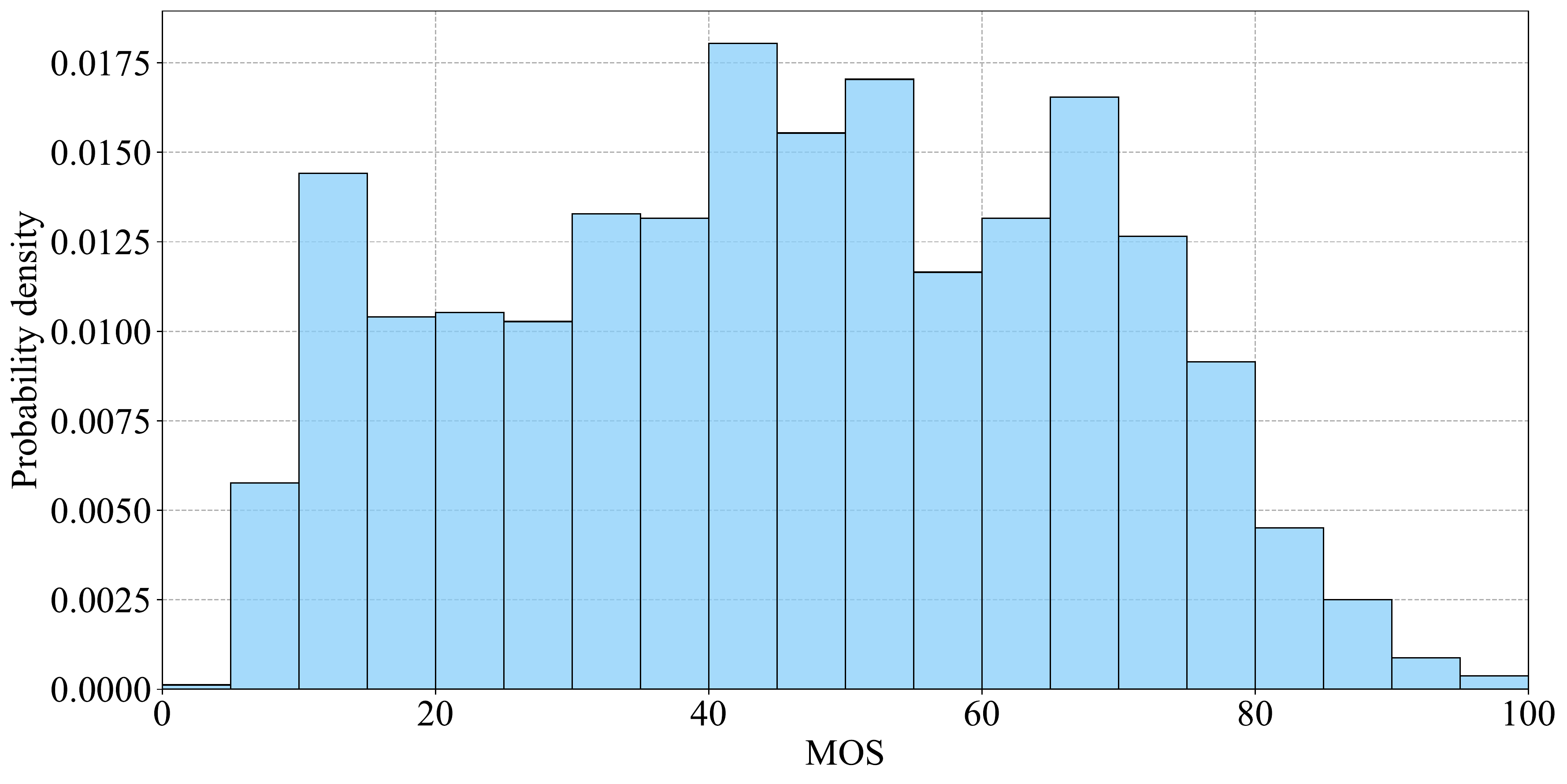}
    \vspace{-0.2cm}
    \caption{The histogram distribution of the MOSs in the proposed database.}
    \label{fig:distribution}
    \vspace{-0.2cm}
\end{figure}

After the subjective experiment,  we calculate the z-scores from the raw ratings as follows:
\begin{equation}
z_{i j}=\frac{r_{i j}-\mu_{i}}{\sigma_{i}},
\end{equation}
where $\mu_{i}=\frac{1}{N_{i}} \sum_{j=1}^{N_{i}} r_{i j}$, $\sigma_{i}=\sqrt{\frac{1}{N_{i}-1} \sum_{j=1}^{N_{i}}\left(r_{i j}-\mu_{i}\right)}$, and $N_i$ is the number of images judged by subject $i$.
The ratings from unreliable subjects according to ITU-R BT.500-13 \cite{bt2002methodology}.
The corresponding z-scores are linearly rescaled to $[0,100]$ and the mean opinion scores (MOSs) are computed by averaging the rescaled z-scores. The MOS distribution is exhibited in Fig. \ref{fig:distribution}, from which we can see that the quality ratings cover most of the quality range.

\vspace{-0.1cm}
\section{Objective Quality Assessment}
\vspace{-0.1cm}
To provide useful guidelines for quality-oriented digital human systems, we propose a simple yet effective full-reference (FR) projection-based method to predict the visual quality of DHH models. The framework of the proposed method is illustrated in Fig. \ref{fig:framework}, which includes the feature extraction, the feature fusion module, and the feature regression module. Only the front projections are used for feature extraction.

\vspace{-0.1cm}
\subsection{Feature Extraction}
\vspace{-0.1cm}
Following the common process of mapping the reference and distorted projections into quality embeddings with DNNs, the pretrained Swin Transformer tiny (ST-t) \cite{liu2021swin} is employed as the feature extraction backbone since it takes up less training and inference resources. Considering the visual information is perceived hierarchically from simple low-level features (e.g., noise and texture) to complex high-level features (e.g., semantic information), we employ the hierarchical structure for feature extraction:

\begin{equation}
\begin{aligned}
     En(x) &= \alpha_{0}(x)\oplus\alpha_{1}(x) \oplus\alpha_{2}(x) \oplus \alpha_{3}(x),\\
      \alpha_{k}(x) &= {\rm{AP}}(S_{k}(x)), k \in \{0,1,2,3\},
\end{aligned} 
\end{equation}
where $S_{k}(x)$ represents the features from the $k$-th stage, ${\rm{AP}}(\cdot)$ stands for the average pooling operation, $\alpha_{k}(x)$ denotes the pooled results from the $k$-th stage, and $\oplus$ indicates the concatenation operation. Given the input reference and distorted front  projections $P_{r}$ and $P_{d}$, the quality-aware features can be obtained as:

\begin{figure}[!tbp]
    \centering
    \includegraphics[width=6.5cm]{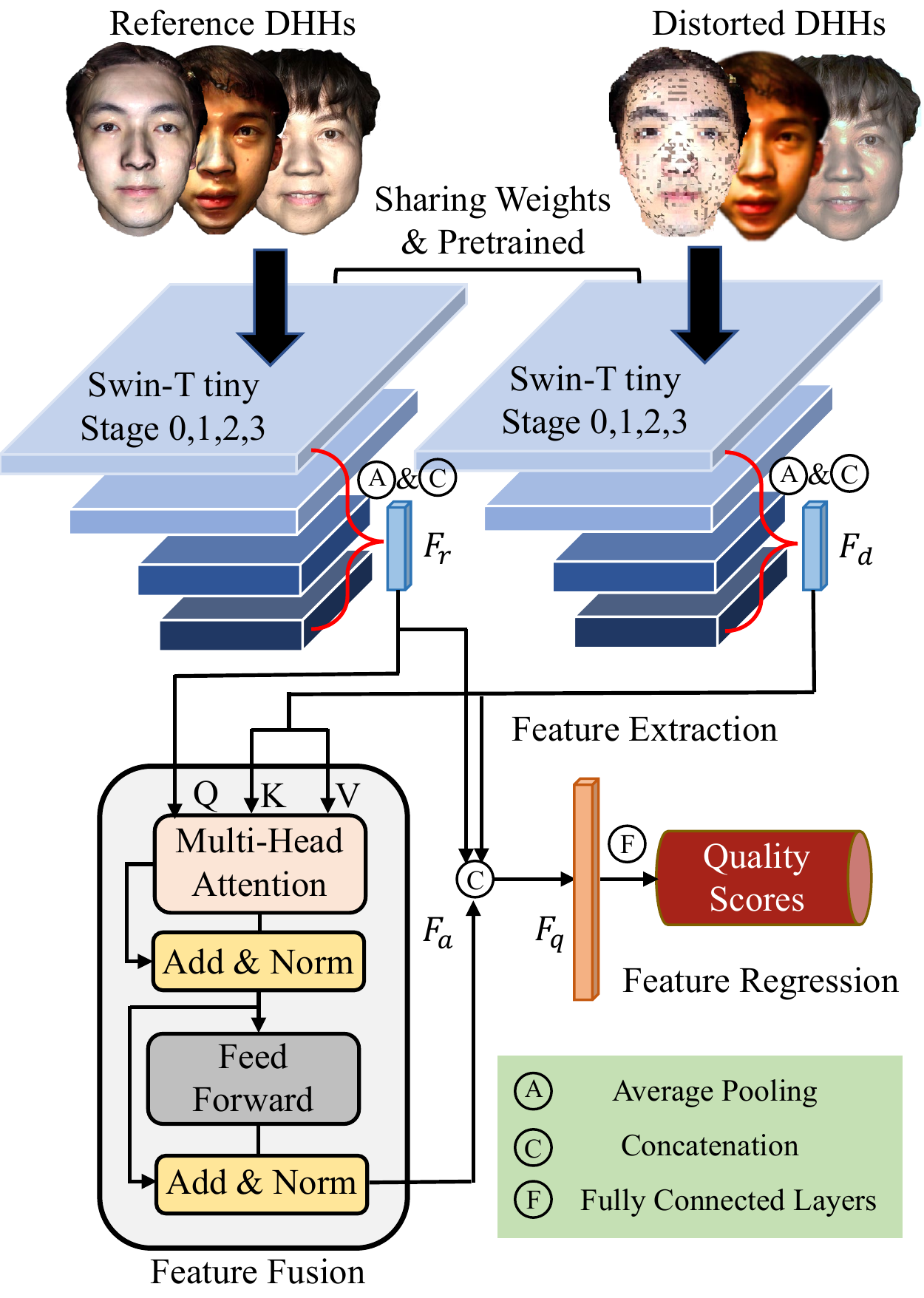}
    \caption{The framework of the proposed method.}
    \label{fig:framework}
    \vspace{-0.2cm}
\end{figure}

\begin{equation}
    F_{r} = En(P_{r}),F_{d} = En(P_{d}), 
\end{equation}
where $En(\cdot)$ is the hierarchical ST-t encoder, $F_{r}$ and $F_{d}$ represent the quality-aware embeddings for the input reference and distorted front  projections.

\vspace{-0.1cm}
\subsection{Feature Fusion}
\vspace{-0.1cm}
To actively relate the quality-aware information between the reference and distorted embeddings, we utilize the multi-head attention operation for feature fusion:

\begin{equation}
\begin{aligned}
    F_{a} = \Gamma(Q& =F_{r},K=F_{d},V=F_{d}), \\
    F_{q} &= F_{r} \oplus F_{d} \oplus F_{a},
\end{aligned} 
\end{equation}
where $\Gamma(\cdot)$ indicates the multi-head attention operation where $F_{r}$ is used to guide the attention learning of $F_{d}$, $F_{a}$ is the fused quality-aware embedding, and $F_{q}$ is the final quality embedding.

\vspace{-0.1cm}
\subsection{Feature Regression}
\vspace{-0.1cm}
With the final quality embedding $F_{q}$, two-stage fully-connected (FC) layers are employed for feature regression:

\begin{equation}
    Q' = \mathbf{FC}(F_{q}),
\end{equation}
where $Q'$ represents the predicted quality scores and $\mathbf{FC}(\cdot)$ stands for the fully-connected layers. Moreover, the L1 loss is utilized as the loss function:

\begin{equation}
    Loss = \frac{1}{n} \sum_{\eta=1}^{n} |Q_{\eta}-Q'_{\eta}|,
\end{equation}
where $n$ is the size of the mini-batch, $Q_{\eta}$ and $Q'_{\eta}$ are quality labels and predicted quality scores respectively.

\vspace{-0.1cm}
\section{Experiment}
\vspace{-0.1cm}
\subsection{Implementation Details \& Competitors}
\vspace{-0.1cm}
The ST-t backbone is fixed with weights pretrained on the ImageNet \cite{deng2009imagenet} and only the feature fusion and regression modules are trained. The Adam optimizer \cite{kingma2014adam} is used with the initial learning rate set as 1e-4 and the number of training epochs is set as 50. The batch size is set as 32. The 1080$\times$1920 front projections are resized into the resolution of 256$\times$456 and patches with the resolution of 224$\times$224 as the input. We conduct the 5-fold cross validation on the DHHQA database and the average results of the 5 folds are recorded as the final performance. It's worth mentioning that there is no content overlap between the training and testing sets.
Several mainstream FR-IQA methods are validated for comparison as well, which include PSNR, SSIM \cite{ssim}, MS-SSIM \cite{ms-ssim}, GMSD \cite{gmsd}, LPIPS \cite{lpips}, and PieAPP \cite{prashnani2018pieapp}. These methods are validated using the same front projections on the same testing sets as the proposed method and the average results are recorded as the final performance. We also include the performance of some classic FR point cloud quality assessment (PCQA) methods such as p2point\cite{cignoni1998metro}, p2plane\cite{tian2017geometric} and PSNR-yuv\cite{torlig2018novel} through converting the DHHs from textured meshes to point clouds. Additionally, all the quality scores predicted by the IQA methods are processed with five-parameter logistic regression to deal with the scale difference.

\vspace{-0.1cm}
\subsection{Performance Discussion}
\vspace{-0.1cm}
The performance results are shown in Table \ref{tab:performance}, from which we can several inspections: (a) The proposed method achieves the higher performance among all the competitors and is about 0.03 ahead of the second place competitor LPIPS (0.9286 vs. 0.8935) in terms of SRCC, which shows the effectiveness of the proposed method for predicting the visual quality of DHH models;  (b) Comparing the performance of MS-SSIM and SSIM, we can conclude that the multi-scale features can greatly help boost the performance since the MS-SSIM surpasses SSIM by almost 0.12 in terms of SRCC, which also reflects the rationality of the proposed hierarchical ST-t structure; (c) The learning-based methods (LPIPS, PieAPP, and proposed method) are all superior to the handcrafted-based methods. It can be explained that the handcrafted-based IQA methods are developed mainly for natural scene images (NSIs). However, the prior knowledge for perceptual quality differs from NSIs to DHH projections since the projections are artificially rendered with computers rather than captured with cameras. By actively learning the prior knowledge from the DHH projections, the learning-based methods can yield better performance consequently.

\begin{table}[t]\small
\renewcommand\arraystretch{0.9}
  \caption{Performance comparison on the DHHQA database. Best in bold. }
  \vspace{-0.2cm}
  \begin{center}
  \begin{tabular}{c|cccc}
    \toprule
     Model &SRCC  &  PLCC & KRCC & RMSE \\ \hline
    MSE-p2point & 0.2891 & 0.2916 & 0.2359 & 21.0813\\
    MSE-p2plane & 0.2698 & 0.2961 & 0.2250 & 21.0502\\
    PSNR-yuv & 0.1761 & 0.2272 & 0.1369 & 21.4299\\
    PSNR  &0.8347 &0.8371 &0.6405 &11.5822\\
    SSIM  &0.7355 &0.7253 &0.5388 &14.5221\\
    MS-SSIM &0.8557 &0.8396 &0.6653 &11.4953\\
    GMSD  &0.8411 &0.8350 &0.6534 &11.6441\\
    LPIPS &0.8935  &0.8881 &0.7085 &9.7194\\
    PieAPP &0.8769 &0.8723 &0.6857 &10.3552\\
    Proposed &\bf{0.9286} & \bf{0.9320} & \bf{0.7585} & \bf{7.2910} \\
    \bottomrule
  \end{tabular}
  \end{center}
  \label{tab:performance}
  \vspace{-0.7cm}

\end{table}

\vspace{-0.1cm}
\section{Conclusion}
\vspace{-0.1cm}
In this paper, we propose a large-scale digital human head quality assessment database to deal with the issues of digital human quality assessment. 55 reference DHHs are selected and each reference DHH is degraded with 7 types of distortions under 4 levels, which generates a total of 1,540 distorted DHH models. Then a subjective quality assessment experiment is carried out to obtain the MOSs of the distorted DHH models. Afterward, a simple yet effective FR projection-based method is proposed by employing the pretrained Swin Transformer tiny as the backbone. The hierarchical quality-aware features are extracted from the reference and distorted DHHs' front projections and fused with the multi-head attention module. The experimental results show that the proposed method outperforms the mainstream FR-IQA methods, which demonstrates its effectiveness for predicting the visual quality levels of DHHs.

\bibliographystyle{IEEEbib}
\bibliography{icme2021template}

\end{document}